\documentclass{article}

\usepackage{arxiv}
\usepackage{amsmath}
\usepackage[utf8]{inputenc} 
\usepackage[T1]{fontenc}    
\usepackage{hyperref}       
\usepackage{url}            
\usepackage{booktabs}       
\usepackage{amsfonts}       
\usepackage{nicefrac}       
\usepackage{microtype}      
\usepackage{graphicx}
\usepackage{natbib}
\usepackage{placeins}

\title{What Single-Prompt Accuracy Misses: A Multi-Variant Reliability Audit of Language Models}

\author{ \href{https://orcid.org/0000-0003-2427-7920}{\includegraphics[scale=0.06]{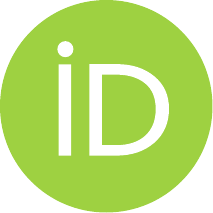}\hspace{1mm}Ranit Karmakar} \\
	Harvard University, 
	Boston, MA 02115 \\
	\And
	\href{https://orcid.org/0009-0001-1814-258X}{\includegraphics[scale=0.06]{orcid.pdf}\hspace{1mm}Jayita Chatterjee} \\
	Boston, MA 02134 \\\
}

\renewcommand{\shorttitle}{Multi-Variant Reliability Audit of Language Models}

\hypersetup{
pdftitle={What Single-Prompt Accuracy Misses: A Multi-Variant Reliability Audit of Language Models},
pdfsubject={cs.CL, cs.AI},
pdfauthor={Ranit Karmakar, Jayita Chatterjee},
pdfkeywords={small language models \and evaluation \and calibration \and prompt robustness \and verbal confidence},
}

\begin{document}
\maketitle

\begin{abstract}
Single-prompt accuracy is the dominant way to benchmark language models, but it can miss reliability failures that matter. We evaluate a 15-model open-weight corpus, with the main reliability analyses focused on 10 instruct models across five classification and reasoning benchmarks under five prompt variants each, measuring accuracy, token-probability calibration, verbal-confidence calibration, verbal parse rate, and prompt-perturbation spread for every (model $\times$ dataset $\times$ variant) cell. We find three broad results. First, where the task instruction sits in the prompt changes classification accuracy by tens of points. Moving the instruction away from a single-sentence input lowers accuracy on 10 of 10 AG News and SST-2 cells across five instruct models, by 0.061 to 0.471, with 95\% bootstrap intervals strictly above zero in every one; the same perturbation leaves the two reasoning benchmarks unchanged. Instruction position is a free parameter that evaluation reports almost never state. Second, confidence signals are fragile. On MMLU-Pro, every primary model verbally reports confidence substantially above both its accuracy and its token-probability confidence on the same rows, and verbal parse rate can collapse for a single model on a single prompt variant. Third, prompt robustness does not track parameter count reliably. Across 10 instruct models, the correlation between model size and prompt-perturbation spread ranges from $-0.244$ to $0.474$ across benchmarks. We separately report two evaluation-side artefacts in our own pipeline: a scorer mismatch that depressed ARC-Challenge accuracy under chain-of-thought prompting, and a calibration-normalisation choice that shifts per-cell ECE by a mean absolute 0.149. Taken together, these results show that reliability conclusions for small language models depend not only on the model being evaluated, but also on how the prompt is written and the pipeline that measures it. We argue that calibration definitions, evaluator logic, verbal parseability, and prompt robustness should be reported explicitly when making reliability claims.
\end{abstract}

\section{Introduction}\label{sec:intro}

Small language models (SLMs) in the 1--8B parameter range are increasingly attractive for local assistants, on-device applications, and multi-agent systems where inference cost, latency, or privacy constraints make larger models impractical. In these settings, benchmark accuracy is necessary but incomplete. A deployed model must not only answer correctly on average; it must also expose usable confidence signals, produce outputs that downstream systems can parse, and remain stable under the prompt variants that users or pipelines actually induce.

Most SLM evaluations still emphasize single-prompt accuracy. This is useful for ranking models, but it hides several reliability questions that matter in deployment. Does the model remain calibrated when the prompt format changes? Does verbal confidence track correctness, or merely sound confident? Can the evaluator reliably extract the intended answer from the generated text? And does robustness improve with parameter count, as practitioners often assume? These questions are difficult to answer from a single accuracy number, and even harder to diagnose when prompt templates, evaluator logic, and calibration definitions are not reported explicitly.

We study these issues through a multi-variant reliability audit of 15 open-weight SLMs from 1B to 8B parameters. We evaluate five classification and reasoning benchmarks under five prompt variants per instance, recording accuracy together with token-probability calibration, verbal-confidence calibration, verbal parse rate, and prompt-perturbation spread for each model--dataset--variant cell. This design lets us observe failures that would be averaged away by model-level or single-prompt reporting.

Our results show that reliability conclusions can depend as much on the evaluation pipeline as on the model being evaluated. First, seemingly reasonable evaluation choices can materially change conclusions: label-set normalization changes token-calibration estimates substantially, and pairing chain-of-thought output with a first-character evaluator produces a large apparent ARC-Challenge accuracy collapse that is largely repaired by changing only the evaluator. Second, confidence signals are fragile: on MMLU-Pro, primary models verbally report confidence far above both accuracy and token-probability confidence, while verbal parseability can fail for a single model on a single prompt variant. Third, prompt robustness is not predicted by parameter count in our 1--8B instruct-model sample; smaller models can be substantially more stable than larger ones under prompt perturbation.

This paper makes three contributions:
\begin{itemize}
    \item We show that instruction position, an unreported free parameter in prompt templates, costs 0.061 to 0.471 accuracy on single-sentence classification across every model we test, with paired bootstrap intervals above zero in all ten cells, and costs nothing on reasoning benchmarks.
    \item We jointly report verbal-confidence direction and verbal parse rate per model--dataset--variant cell, showing that verbal confidence can be both overconfident and selectively unparseable.
    \item We measure prompt-perturbation spread across 10 instruct models and show that model size is a poor proxy for robustness in this range.
    \item We document two evaluation-side artefacts in our own pipeline---a scorer mismatch
    and a calibration-normalisation choice---that changed our results without changing any model.
\end{itemize}

Together, these findings argue for reporting SLM reliability as a deployment-readiness profile rather than a single benchmark score: calibration definitions, evaluator logic, verbal parseability, and prompt robustness should be made explicit whenever reliability claims are made.

\section{Related Work}\label{sec:related}

\paragraph{Calibration and verbal confidence in language models:}
Expected Calibration Error (ECE) is a standard summary of probabilistic calibration, building on binned reliability estimates~\cite{naeini2015obtaining,guo2017calibration}. In language models, calibration can be measured from token probabilities, from generated verbal confidence, or from post-hoc transformations of either signal. Prior work has shown that larger RLHF-tuned models can sometimes verbalize useful uncertainty estimates~\cite{kadavath2022language,tian2023just}, while other studies find systematic verbal overconfidence, especially on harder tasks~\cite{xiong2023can}. Related work has also trained models to express uncertainty directly in natural language~\cite{lin2022teaching}. Our work focuses on the smaller 1--8B open-weight regime and measures token confidence, verbal confidence, and verbal parseability jointly at the level of each model--dataset--prompt-variant cell, rather than treating calibration as a single model-level property.

\paragraph{Prompt sensitivity and evaluation design:}
Language-model evaluations are sensitive to prompt wording, formatting, and in-context examples. Prior work has shown that prompt-based models may not use instructions in semantically literal ways~\cite{webson2022prompt}, that few-shot prompts can introduce output-distribution biases~\cite{zhao2021calibrate}, and that semantically similar or superficially reformatted prompts can produce large accuracy differences~\cite{mizrahi2024state,sclar2023quantifying}. Multi-dimensional evaluation frameworks such as HELM~\cite{liang2022holistic}, BIG-bench~\cite{srivastava2023beyond}, and the EleutherAI LM Evaluation Harness~\cite{gao2021framework} have made broad, standardized evaluation more practical, with HELM explicitly including robustness and calibration as evaluation axes. However, most prompt-sensitivity studies emphasize accuracy variance, while most benchmark harnesses fix a canonical prompt and evaluator. Our work studies how prompt perturbations interact not only with accuracy, but also with calibration definitions, verbal-confidence parsing, and evaluator logic.

\paragraph{Evaluation of small open-weight models:}
Recent open-weight SLM releases, including Phi, Llama, Gemma, Qwen, SmolLM, and DeepSeek-R1 distillations, are accompanied by extensive benchmark reports~\cite{abdin2024phi,grattafiori2024llama,Kamath2025Gemma3T,yang2025qwen3,allal2025smollm2,guo2025deepseek}. These reports are valuable for comparing capabilities across standard tasks, but they primarily emphasize single-prompt accuracy, instruction following, safety, and efficiency. Calibration metrics, prompt-variant robustness, verbal-confidence behavior, and evaluator compatibility are rarely co-reported for the same models on the same examples. This paper complements model-release benchmarks by auditing reliability-side quantities across 15 open-weight SLMs, five benchmarks, and five prompt variants, showing that reliability conclusions can change with the evaluation pipeline itself.

\section{Methods}\label{sec:methods}

We study how evaluation design changes the reliability conclusions one would draw about small language models. Our evaluation covers 15 open-weight models in the 1--8B parameter range, with the main claims carried by a primary set of five instruct models and supplementary models used to widen the robustness analysis and exploratory appendices. We evaluate the models on five primary benchmarks: three classification tasks (SST-2, MNLI, AG News) and two reasoning tasks (ARC-Challenge, MMLU-Pro). For each (model, dataset, prompt variant) cell, we evaluate 500 examples.

To make evaluation choices visible rather than incidental, we run each instance under five prompt variants: \textit{surface\_paraphrase}, \textit{instruction\_reorder}, \textit{fewshot\_3}, \textit{format\_change}, and \textit{implicit\_framing}. These variants perturb phrasing, instruction order, in-context conditioning, output format, and framing while keeping the underlying task fixed. Exact templates are provided in the appendix and repository.

For each cell, we record accuracy, token-probability calibration, verbal-confidence calibration, verbal parse rate (VPR), and prompt-perturbation spread. For multiple-choice tasks, accuracy is computed by \textit{is\_correct\_mc}, which marks an example correct if the generation, or its last non-empty line, begins with the gold choice letter; token confidence is computed separately via \textit{extract\_pred\_letter}, which reverse-scans lines for a standalone choice letter when the first character is not a valid label. These are two different definitions of the model's prediction, and they diverge on chain-of-thought outputs (Appendix~A.3). Token confidence is computed from a dedicated \textit{max\_tokens = 1} generation with \textit{logprobs = 200}; our main token-calibration metric uses the label-set-normalised probability mass over observed valid answer letters. Verbal confidence is collected in a follow-up call under two structurally distinct elicitation phrasings. The primary phrasing, used for all main-text claims unless otherwise noted, requests a decimal probability between 0 and 1; the replication phrasing requests a percentage on a 0--100 scale and differs from the primary on scale, mood, and output format. Exact prompt strings for both are released with the public code repository. VPR is the fraction of responses from which a numerical confidence can be parsed, and we use a VPR threshold of 0.80 when aggregating verbal-calibration results. Prompt-perturbation spread is defined as the max--min accuracy across the four \textit{non-format\_change} variants, because \textit{format\_change} on ARC-Challenge reflects an evaluator-pipeline interaction rather than ordinary robustness. All ECE values use equal-width binning with \textit{n\_bins = 10}.

All runs use vLLM with models loaded in bfloat16 on a single NVIDIA L4 GPU (23.7 GB VRAM), with random seed 42 throughout. For the reasoning \textit{format\_change} variant, generation length is extended to allow chain-of-thought output; the verbal-confidence follow-up uses a separate generation budget. Full prompt templates, software versions, and raw per-cell outputs are released with the public code repository.

\section{Findings}\label{sec:findings}

\subsection{Finding 1 --- Where the instruction sits changes classification accuracy by tens of points}

Prompt templates differ in a way that papers rarely report: whether the task instruction
precedes the input or follows it. Both orders are natural, both appear in released harnesses,
and neither is obviously correct. On single-sentence classification, the choice costs every
model we tested a large amount of accuracy.

\begin{figure}[ht!]
    \centering
    \includegraphics[width=0.86\linewidth]{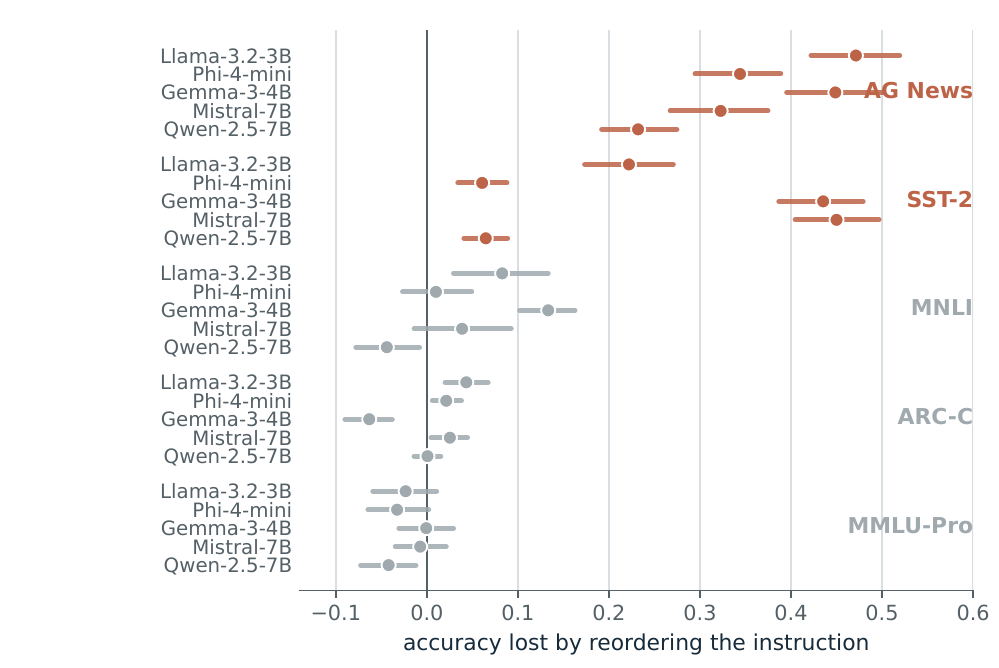}
    \caption{Accuracy lost by moving the task instruction away from the input, paired at the
    example level against the mean of the three variants that keep it adjacent. Bars are 95\%
    percentile-bootstrap intervals. On the two single-sentence classification benchmarks the
    interval excludes zero for every model; on the reasoning benchmarks the effect is absent.}
    \label{fig:order}
\end{figure}

Our \textit{instruction\_reorder} variant places the instruction and label list ahead of the
input and asks for the answer after it; the three other non-chain-of-thought variants keep the
instruction adjacent to the question. Paired at the example level across the five primary
instruct models, reordering lowers accuracy on \textbf{10 of 10} AG News and SST-2 cells, with
95\% percentile-bootstrap intervals strictly above zero in every one. The cost runs from
$0.061$ to $0.471$ accuracy; the largest are Llama-3.2-3B on AG News ($0.471$, CI
$[0.422, 0.519]$) and Mistral-7B on SST-2 ($0.450$, CI $[0.405, 0.497]$), shown in
Figure~\ref{fig:order} and Table~\ref{tab:order}. Across all 25 primary cells, reordering is
never the best of the four variants.

The effect is specific to short inputs. On the two reasoning benchmarks the same perturbation
moves accuracy between $-0.063$ and $+0.043$, and on MNLI, a sentence-pair task, only two of
five models show an interval above zero. One reading is that when the input is a single
sentence, separating the instruction from it leaves the text ambiguous between something to
classify and something to continue, while a longer question carries its own framing. We do not
test that mechanism directly.

This matters because instruction position is a free parameter that evaluation reports almost
never state. Two groups running the same benchmark, on the same model, with the same decoder,
can differ by up to $0.47$ accuracy on AG News through this choice alone, with neither doing
anything unusual. Per-cell values and intervals for all 25 cells are in the supplement.

\begin{table}[ht!]
\centering
\caption{\textbf{Instruction reordering on single-sentence classification.} Accuracy with the
instruction adjacent to the input (mean of three variants) against accuracy with it moved, and
the paired difference with a 95\% percentile-bootstrap interval. $n=500$ per cell.}
\begin{tabular}{llcccc}
\toprule
\textbf{Benchmark} & \textbf{Model} & \textbf{Adjacent} & \textbf{Reordered} & \textbf{Drop} & \textbf{95\% CI} \\
\midrule
AG News & Llama-3.2-3B & 0.747 & 0.276 & 0.471 & [0.422, 0.519] \\
 & Phi-4-mini & 0.818 & 0.474 & 0.344 & [0.295, 0.389] \\
 & Gemma-3-4B & 0.787 & 0.338 & 0.449 & [0.395, 0.501] \\
 & Mistral-7B & 0.739 & 0.416 & 0.323 & [0.267, 0.375] \\
 & Qwen-2.5-7B & 0.892 & 0.660 & 0.232 & [0.192, 0.275] \\
\midrule
SST-2 & Llama-3.2-3B & 0.842 & 0.620 & 0.222 & [0.173, 0.271] \\
 & Phi-4-mini & 0.921 & 0.860 & 0.061 & [0.034, 0.088] \\
 & Gemma-3-4B & 0.923 & 0.488 & 0.435 & [0.387, 0.479] \\
 & Mistral-7B & 0.884 & 0.434 & 0.450 & [0.405, 0.497] \\
 & Qwen-2.5-7B & 0.897 & 0.832 & 0.065 & [0.041, 0.089] \\
\bottomrule
\end{tabular}
\label{tab:order}
\end{table}

\subsection{Finding 2 --- Confidence signals are fragile, and accuracy alone will not reveal that}

We next measure verbal confidence jointly with token-probability confidence and accuracy on the same rows. This lets us ask two separate questions: whether verbal confidence is directionally reliable when produced, and whether it is parseable at all.

On MMLU-Pro (10-choice; primary-model accuracy 0.16--0.29 on the full sample), every primary instruct model overstates its confidence verbally. Averaged over variants passing the $VPR \geq 0.80$ inclusion threshold, mean verbal confidence is 60--78 percentage points above accuracy and 25--60 points above mean token-probability confidence on the same rows. \textit{ECE\_gap} ($=$ \textit{ECE\_verbal} $-$ \textit{ECE\_token}) is positive for every model, ranging from $+0.286$ to $+0.582$ in Table~\ref{tab:mmlu_ece_components}. Per-cell signed-quantity bootstrap CIs are strictly above zero across all 22 contributing cells. The pattern also survives a structurally different elicitation phrasing: under a percentage-scale interrogative prompt, the sign of both signed measures is preserved on every model, and four of five models shift by less than 0.10.

\begin{table} [ht!]
\centering
\caption{\textbf{MMLU-Pro ECE components per primary instruct model} (means over variants with VPR $\geq 0.80$; $n = 2$--$3$ valid variants per cell).}
\begin{tabular}{lcccc}
\toprule
\textbf{Model} & \textbf{Accuracy} 
& \textbf{ECE (token-level)} 
& \textbf{ECE (verbalized)} 
& \textbf{$\Delta$ECE} \\
\midrule
Gemma-3-4B   & 0.157 & 0.497 & 0.783 & +0.286 \\
Qwen-2.5-7B  & 0.259 & 0.315 & 0.673 & +0.358 \\
Phi-4-mini   & 0.294 & 0.192 & 0.608 & +0.416 \\
Llama-3.2-3B & 0.227 & 0.085 & 0.625 & +0.540 \\
Mistral-7B   & 0.275 & 0.125 & 0.707 & +0.582 \\
\bottomrule
\end{tabular}
\label{tab:mmlu_ece_components}
\end{table}

The overconfidence is task-dependent rather than a single global model property. On easier benchmarks (SST-2, AG News, MNLI, ARC-Challenge), per-model \textit{ECE\_gap} is typically near zero or negative and mixed in sign, with individual cells roughly in the range $-0.32$ to $+0.30$. The strongest failures occur when the model is genuinely uncertain---low accuracy and dispersed first-token probability---but still emits high natural-language confidence.

\begin{figure}[ht!]
    \centering
    \includegraphics[width=\linewidth]{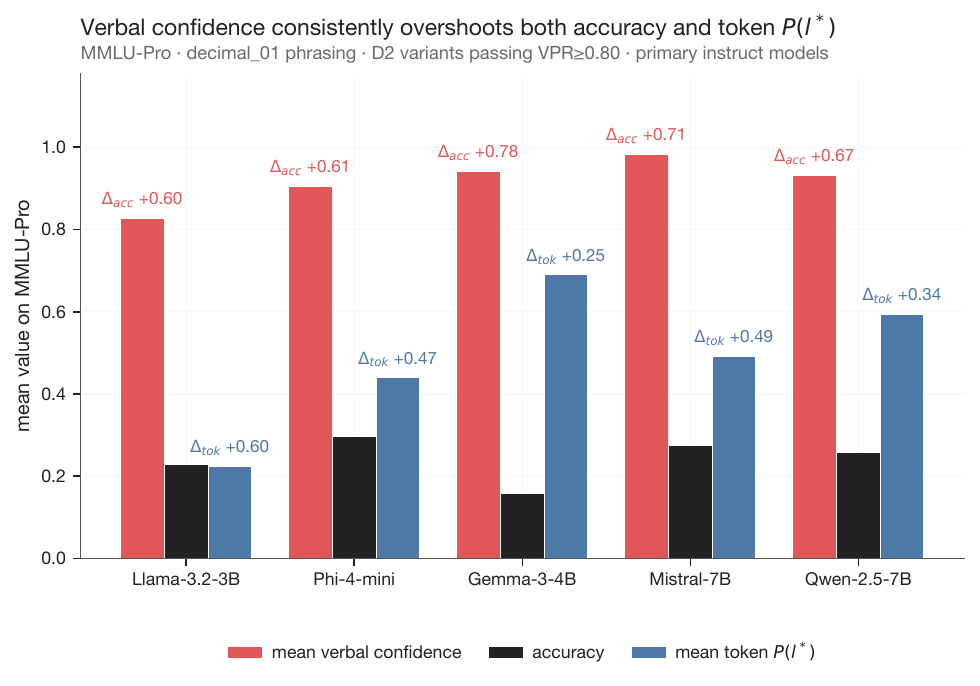}
    \caption{Verbal confidence systematically overstates both accuracy and token-probability confidence on MMLU-Pro. The gap remains strongly positive across all primary instruct models on variants meeting the verbal parse rate inclusion threshold ($VPR \geq 0.80$).}
    \label{fig:find2}
\end{figure}

A separate failure appears before calibration is even computed: the free-text confidence must be parseable. Mistral-7B-Instruct under \textit{surface\_paraphrase} fails the 0.80 VPR threshold on every dataset, with three of five datasets near zero (SST-2 0.002, MNLI 0.046, AG News 0.006). On the other three non-\textit{format\_change} variants, Mistral's VPR ranges from 0.29 to 1.00 and clears the threshold in 12 of 15 cells. The collapse is therefore variant-specific, not model-level: averaging these cells into one VPR scalar would yield a seemingly moderate 0.6--0.7 while silently excluding roughly 20--25\% of verbal outputs from \textit{ECE\_verbal}. The cross-dataset variation pattern survives the alternative percentage-scale elicitation, although absolute magnitudes shift; below-threshold cells remain, but their exact identities are partly phrasing-dependent. Phi-4-mini also falls below threshold on one cell (MMLU-Pro \textit{surface\_paraphrase}, VPR = 0.756), showing that parse failures are not isolated to one family (Figure~\ref{fig:find2}).

These results separate two notions often collapsed into the phrase ``verbal confidence works.'' One asks whether the model reports calibrated probabilities when it produces parseable confidence; the other asks whether it produces parseable confidence at all. Either can fail independently. For this reason, we report VPR per (model $\times$ dataset $\times$ variant) cell, use an explicit inclusion threshold, and pair \textit{ECE\_gap} with signed quantities such as verbal confidence minus accuracy. Per-cell signed-overconfidence CIs, reliability diagrams, the full Mistral VPR matrix, and cross-phrasing comparisons are provided in the supplement.

\subsection{Finding 3 --- Prompt robustness is real, but parameter count is a poor proxy for it}

Prompt-perturbation spread is the gap between a model's best and worst accuracy across surface, ordering, in-context, and framing variants on the same benchmark. If parameter count were a reliable proxy for robustness, spread should decrease with model size. In our 10-instruct-model sample (1.0--7.6B), it does not.

Spearman correlations between size and spread are weak and inconsistent: $+0.261$ ($p = 0.466$) on SST-2, $+0.474$ on MNLI, $-0.244$ on AG News, $+0.015$ on ARC-Challenge, and $+0.304$ on MMLU-Pro, with all $p > 0.10$. None reaches 0.50 in absolute value. The SST-2 scatter makes the non-monotonicity concrete: Llama-3.2-1B has spread 0.202, while the 7B DeepSeek-R1-distill-Qwen point has spread 0.916; meanwhile, mid-sized models such as Phi-4-mini at 3.8B (spread 0.084) and Qwen-2.5-3B at 3.1B (spread 0.028) are among the most robust. The cleanest pairwise contrast is Phi-4-mini at 3.8B with SST-2 spread 0.084 [95\% CI 0.050, 0.120] versus Mistral-7B at 7.2B with spread 0.500 [CI 0.448, 0.546], a roughly six-fold gap with non-overlapping CIs.

\begin{figure}[ht!]
    \centering
    \includegraphics[width=\linewidth]{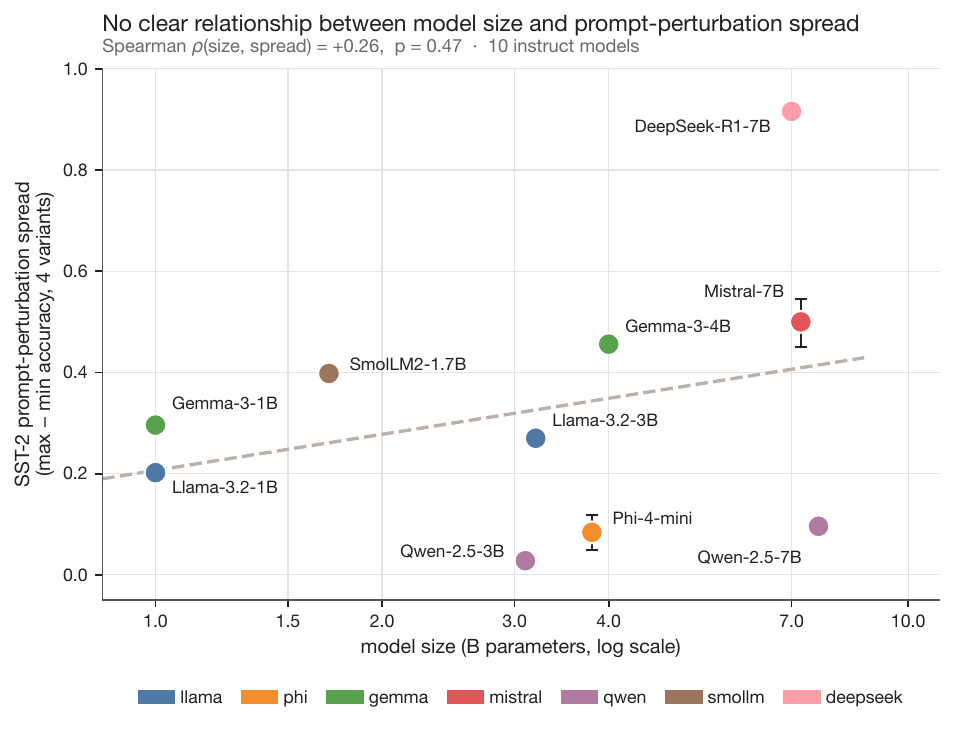}
    \caption{Prompt-perturbation spread on SST-2 vs. model parameter count for 10 instruct models. Spread does not monotonically decrease with model size (Spearman $\rho = +0.26$, $p = 0.47$), indicating that parameter count is a poor proxy for robustness. Within-family clustering is also visible.}
    \label{fig:find3}
\end{figure}

\begin{table*}[ht!]
\centering
\caption{\textbf{Prompt-perturbation spread for all 10 instruct models} (max $-$ min accuracy over four non-\texttt{format\_change} variants). Bold marks the two lowest and two highest values per column.}
\begin{tabular}{lccccccc}
\toprule
\textbf{Model} & \textbf{Size} & \textbf{Family} & \textbf{SST-2} & \textbf{MNLI} & \textbf{AG News} & \textbf{ARC-C} & \textbf{MMLU-Pro} \\
\midrule
Llama-3.2-1B        & 1.0B & llama    & 0.202 & 0.110 & 0.414 & 0.098 & \textbf{0.036} \\
Gemma-3-1B          & 1.0B & gemma    & 0.296 & 0.160 & 0.464 & 0.176 & 0.144 \\
SmolLM2-1.7B        & 1.7B & smollm   & 0.398 & \textbf{0.008} & 0.456 & 0.056 & 0.110 \\
Qwen-2.5-3B         & 3.1B & qwen     & \textbf{0.028} & 0.402 & \textbf{0.610} & 0.054 & 0.084 \\
Llama-3.2-3B        & 3.2B & llama    & 0.270 & 0.110 & 0.536 & 0.098 & \textbf{0.064} \\
Phi-4-mini          & 3.8B & phi      & \textbf{0.084} & \textbf{0.030} & \textbf{0.374} & \textbf{0.046} & 0.106 \\
Gemma-3-4B          & 4.0B & gemma    & 0.456 & 0.254 & 0.494 & \textbf{0.370} & 0.156 \\
DeepSeek-R1-7B & 7.0B & deepseek & \textbf{0.916} & \textbf{0.538} & \textbf{0.754} & \textbf{0.568} & \textbf{0.260} \\
Mistral-7B          & 7.2B & mistral  & \textbf{0.500} & 0.254 & \textbf{0.374} & 0.104 & \textbf{0.014} \\
Qwen-2.5-7B         & 7.6B & qwen     & 0.096 & 0.218 & 0.236 & \textbf{0.048} & \textbf{0.188} \\
\bottomrule
\end{tabular}
\label{tab:spread_all_models}
\end{table*}

The pattern is more structured within families than across size alone. Both Qwen models are among the most robust on SST-2 (spread 0.028 and 0.096); both Gemma models are among the most brittle on ARC-Challenge (spread 0.176 and 0.370); and the reasoning-distilled DeepSeek-R1-distill-Qwen point is the high-spread outlier on classification. We treat this as a candidate organizing pattern, not a causal claim about family or training recipe: with 1--2 models per family and a single distilled-reasoning point, the sample cannot separate family effects from recipe effects or convenience-of-sample effects.

The practical conclusion is narrow: parameter count should not be used as a stand-in for prompt robustness in the 1--8B range. For deployments where prompt format is not tightly controlled, prompt-perturbation spread should be reported directly alongside accuracy. Full per-model spread values, family annotations, and per-benchmark correlation tests are provided in the supplement.

\subsection{Two evaluation-side artefacts in our own pipeline}

Two further effects sit upstream of the numbers above: the evaluator that turns generated text into a judged label, and the definition of the calibration metric. We report them because they changed our own results, not because we believe either is standard practice. The first in particular is a defect we found in our own harness.

\begin{figure}[ht!]
    \centering
    \includegraphics[width=\linewidth]{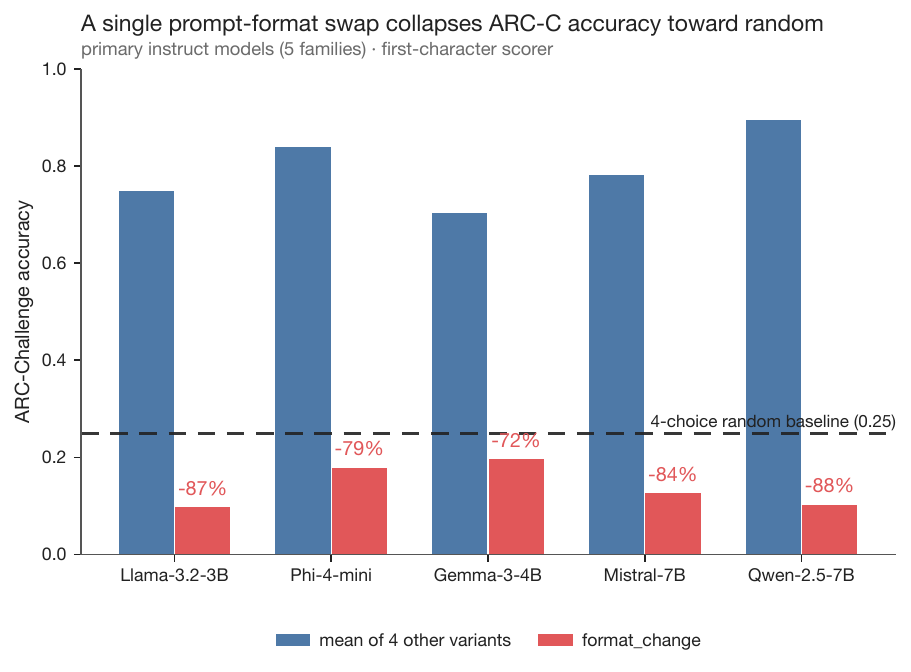}
    \caption{ARC-Challenge accuracy collapse under chain-of-thought elicitation (\textit{format\_change}) when scored by the first-character/last-line rule used in our runs, compared to the mean of four other prompt variants. The relative drop is severe across all five primary instruct models, all of which fall below the 4-choice random baseline (0.25). Under the reverse-scan rule described in Appendix~A.3 the same generations yield 0.472--0.712 and none falls below the baseline.}
    \label{fig:find1}
\end{figure}

On ARC-Challenge, accuracy averaged over the four non-chain-of-thought variants ranges from 0.705 to 0.897 across the five primary models. Switching to a \textit{format\_change} variant that asks for chain-of-thought reasoning collapses accuracy to $0.098$--$0.198$ under the first-character/last-line rule used in our runs (\textit{is\_correct\_mc}), a $72$--$88\%$ relative drop in every training family, with paired bootstrap CIs strictly non-overlapping with zero. Scored instead by the reverse-scan rule described in our appendix (\textit{extract\_pred\_letter}), the same generations yield $0.472$--$0.712$; measured against that rule's own non-chain-of-thought baseline, the drop is $5.9$--$43.0\%$. All five models fall below the 0.25 random baseline for 4-choice MCQ under the first-character/last-line rule; none does under the reverse-scan rule, as shown in Figure~\ref{fig:find1}.

The collapse is an evaluator artefact. The scorer used in our runs marks an example correct only if the generation, or its last non-empty line, begins with the gold letter; under chain-of-thought prompting the generation opens with a reasoning word (``The...'', ``Looking...'', ``Let...'', ``First...'') and the answer letter is embedded in prose rather than beginning the final line, so the answer is marked wrong regardless of what the model concludes later. Two repairs applied to the \textit{same} generations confirm this. A post-hoc regex re-parse that searches for answer markers such as \texttt{Final answer: X}, \texttt{the correct answer is X}, and \texttt{Answer: X} recovers 93.8\% of the gap in aggregate (ratio of means; see Table~\ref{tab:gap_recovery}). A constrained-decoding pass that appends \texttt{\textbackslash n\textbackslash nFinal answer:} to the existing CoT and generates one letter under vLLM \textit{guided\_choice = A/B/C/D} recovers 102.0\% on the same basis. Both repairs leave fewer than 3\% of previously-correct predictions wrong ($\leq 14$ of 500 per model). The two repair paths diverge substantively only on Qwen-2.5-7B (76.4\% vs. 99.9\% recovery), where pathological repetition under the \textit{max\_tokens = 256} cap confuses the regex but not the constrained decoder. Overall, changing only the evaluator moves ARC-Challenge accuracy by 61--67 percentage points on average across the five models as shown in Table~\ref{tab:gap_recovery}.

\begin{table}[ht!]
\centering
\caption{\textbf{Recovery of ARC-Challenge performance under two repair paths} (regex re-parse and constrained decoding) applied to the same chain-of-thought generations. Per-model \textbf{Gap rec.} is that model's own ratio; the \textbf{Mean} row reports a ratio of means computed from the column means in the same row, so both repair paths are summarised under a single aggregation.}
\begin{tabular}{lcccccc}
\toprule
\textbf{Model} & \textbf{Non-fc mean} & \textbf{First-char} 
& \multicolumn{2}{c}{\textbf{Path 1}} 
& \multicolumn{2}{c}{\textbf{Path 2}} \\
\cmidrule(lr){4-5} \cmidrule(lr){6-7}
 &  &  & \textbf{(regex)} & \textbf{Gap rec.} 
 & \textbf{(constrained)} & \textbf{Gap rec.} \\
\midrule
Llama-3.2-3B & 0.751 & 0.098 & 0.712 & 94.0\% & 0.732 & 97.1\% \\
Phi-4-mini   & 0.842 & 0.180 & 0.848 & 100.9\% & 0.862 & 103.0\% \\
Gemma-3-4B   & 0.705 & 0.198 & 0.758 & 110.5\% & 0.786 & 116.0\% \\
Mistral-7B   & 0.783 & 0.128 & 0.748 & 94.7\% & 0.766 & 97.4\% \\
Qwen-2.5-7B  & 0.897 & 0.104 & 0.710 & 76.4\% & 0.896 & 99.9\% \\
\midrule
Mean         & 0.796 & 0.142 & 0.755 & 93.8\% & 0.808 & 102.0\% \\
\bottomrule
\end{tabular}
\label{tab:gap_recovery}
\end{table}

A second evaluation-side choice appears inside the calibration metric. \textit{ECE\_token} can be computed from the raw first-token probability of the predicted letter, $P(l^*)$, or from the label-set-normalised probability, $P(l^*) / \sum_{l \in \mathcal{L}} P(l)$, where $\mathcal{L}$ is the set of valid labels. Both are first-token quantities and both are commonly reported under the same name, but they answer different questions. Across 125 primary cells (5 models $\times$ 5 datasets $\times$ 5 variants), the per-cell mean absolute ECE difference between the two definitions is 0.149 [95\% bootstrap CI 0.118, 0.184]. 54.4\% of cells shift by more than 0.05; 24.0\% shift by more than 0.20; the largest shift is 0.825 on Mistral-7B-Instruct's SST-2 \textit{format\_change} cell (0.891 unnormalised, 0.066 normalised). Shifts are largest on \textit{fewshot\_3} and \textit{format\_change}, where probability mass often spreads across in-context-example letters or reasoning words before the answer label. These figures pair confidence in the letter identified by \textit{extract\_pred\_letter} with correctness as judged by \textit{is\_correct\_mc} (Appendix~A.3). Recomputed with a single prediction definition driving both, the per-cell mean absolute shift is 0.114, with 32.0\% of cells shifting by more than 0.05 and 18.4\% by more than 0.20; the effect is smaller than the figures above but does not vanish.

Both failures sit upstream of the model. A leaderboard number alone does not reveal whether calibration was label-set-normalised or whether the evaluator was compatible with the prompt. We therefore report calibration definitions and evaluator logic explicitly, and retain full generations so results can be re-scored under alternative evaluators. Per-variant calibration shifts, top-200 logprob coverage, and bootstrap details are provided in the supplement.

\section{Recommendations}\label{sec:rec}

The three lessons in \S~\ref{sec:findings} suggest a concrete reporting checklist for SLM reliability audits. Each item is computable from per-cell quantities already produced by a multi-variant evaluation.

\textbf{Disclose \textit{ECE\_token} normalisation:} State whether token-probability ECE is computed from raw first-token probability or from the label-set-normalised form. Prefer the normalised form for cross-paper and cross-cardinality comparability. Where space permits, report the per-cell shift under the alternate definition on at least a representative subset of cells.

\textbf{State where the instruction sits relative to the input:} Give the template verbatim,
and on single-sentence classification report at least one alternative ordering. In our corpus
this one choice moves accuracy by up to 0.47, more than any model substitution we make.

\textbf{Publish evaluator logic alongside the prompt template:} Specify the exact rule that turns generated text into a scored label: first-character match, regex with fallback, constrained decoding, or another rule. Where the same prompt is scored more than one way, report all evaluator--prompt pairs.

\textbf{Retain full generation text:} Storing raw generations lets any reported accuracy number be re-scored under an alternative evaluator if the original evaluator later turns out to interact badly with a prompt format.

\textbf{Report verbal parse rate per cell, not as a model scalar:} Apply an explicit inclusion threshold for \textit{ECE\_verbal} means and footnote the excluded cells. Alongside any aggregate calibration gap, report at least one signed quantity, such as mean verbal confidence minus accuracy on the rows where verbal confidence parses.

\textbf{Replicate verbal-confidence claims under at least two elicitation phrasings:} Use phrasings that differ in scale, mood, and output format. Single-phrasing results should not be read as model properties; in our setup, the qualitative claims survive this test, but absolute magnitudes do not.

\textbf{Report prompt-perturbation spread alongside accuracy when comparing robustness:} Use a fixed variant set defined in the paper. In our 10-model sample, parameter count does not order models by spread, so we recommend reporting spread directly rather than treating size as a stand-in for robustness.

Together, these items require little additional reporting once a multi-variant evaluation has been run. The information they add for a downstream practitioner is the difference between a leaderboard and a deployment-readiness profile.

\section{Limitations}\label{sec:limit}

This study is observational. The findings are properties of a fixed evaluation corpus: 15 open-weight models, five English benchmarks, five prompt variants, two verbal-confidence elicitation phrasings, and one inference stack. The results support claims about how evaluation choices affect reliability conclusions in this setting; they do not establish causal claims about training recipes, model families, or scaling behavior.

\textbf{Model coverage:} The model sample is uneven across families, with only one or two models per family in most cases. The primary claims are evaluated on five instruct models, while the robustness analysis uses 10 instruct models. Findings that appear consistent across families, especially the evaluator failure and MMLU-Pro overconfidence, should be replicated on broader model sets and outside the 1.0--7.6B range. The within-family patterns in prompt robustness are descriptive only and cannot be separated from training-recipe or sample-selection effects.

\textbf{Task and language scope:} The main benchmarks are English classification or multiple-choice reasoning tasks. We do not evaluate open-ended generation, long-context reasoning, code generation, multilingual tasks, or interactive deployment settings. In particular, the verbal-overconfidence result is strongest on high-cardinality MCQ reasoning; whether the same pattern holds for open-ended generation remains open.

\textbf{Evaluation and inference choices:} All calibration results use equal-width ECE binning with $n_{\text{bins}}=10$, and token calibration uses the constrained-choice normalisation $P(l^*)/\sum_{l \in \mathcal{L}}P(l)$. Other binning schemes or confidence definitions may change per-cell ECE values. Runs use vLLM with bfloat16 on a single NVIDIA L4 GPU, so first-token logprob distributions and efficiency measurements may differ under other inference engines or hardware.

\textbf{Verbal-confidence elicitation:} We test two structurally different confidence phrasings, and the main qualitative claims survive both. However, two phrasings do not exhaust the space of possible elicitations. Chain-of-thought self-assessment, categorical confidence scales, third-person framings, or task-specific calibration prompts could produce different absolute magnitudes.

These limitations define the scope of the claims: within this controlled corpus, reliability conclusions depend not only on the model, but also on calibration definitions, evaluator logic, parseability, and prompt perturbations.

\section{Conclusion}\label{sec:conc}

Single-prompt accuracy gives an incomplete picture of small language model reliability. Across 15 open-weight 1--8B models, five benchmarks, and five prompt variants, we find that reliability conclusions can change with evaluation choices that are often treated as incidental: how token confidence is normalised, how generated text is scored, whether verbal confidence can be parsed, and how robustness is measured across prompt variants.

The central implication is that SLM evaluation should report the pipeline, not only the score. Calibration definitions, evaluator logic, raw generations, verbal parseability, and prompt-perturbation spread are not bookkeeping details; they determine whether a reported reliability claim is reproducible, auditable, and useful for deployment. If SLM benchmarks are meant to guide real model selection, they should describe not only which model performed best, but also which evaluation assumptions made that conclusion possible.

\bibliographystyle{plainnat}
\bibliography{references}

\newpage
\appendix

\section*{Appendix}
\addcontentsline{toc}{section}{Appendix}

This appendix consolidates supporting evidence for the three findings in
the main text. Appendix~\ref{app:repro} documents the experimental
protocol and implementation details. Appendix~\ref{app:prompts}
reproduces the prompt templates and verbal-confidence elicitation
strings. Appendix~\ref{app:finding1} provides additional evidence for
Finding~1 (ARC-Challenge evaluator repairs and calibration-normalization
sensitivity). Appendix~\ref{app:finding2} provides additional evidence
for Finding~2 (verbal parse rate, the Mistral parse-rate matrix,
cross-phrasing robustness, and reliability diagrams).
Appendix~\ref{app:finding3} provides additional evidence for Finding~3
(size--spread correlations and the bootstrap procedure).

\section{Experimental details}
\label{app:repro}

\subsection{Models and datasets}

We evaluate fifteen open-weight models in the 1.0--7.6B parameter range. 
The main quantitative analyses use ten instruct models. Within these, the 
primary instruct set of five models carries the main-text claims about 
evaluator failure, calibration-normalization sensitivity, and verbal-confidence 
behavior: Llama-3.2-3B-Instruct, Phi-4-mini-Instruct, Gemma-3-4B-IT, 
Mistral-7B-Instruct, and Qwen-2.5-7B-Instruct. A supplementary instruct 
set of five models widens the robustness analysis in Finding 3: 
Llama-3.2-1B-Instruct, Gemma-3-1B-IT, SmolLM2-1.7B-Instruct, 
Qwen-2.5-3B-Instruct, and DeepSeek-R1-distill-Qwen-7B.

The remaining five models are base-model counterparts used only for 
supplementary exploratory analyses and released artifacts; they do not 
contribute to the main-text claims. This separation lets the paper report 
a 15-model evaluation corpus while keeping the main reliability claims 
focused on the 10 instruct models most relevant to deployment-style 
evaluation.

\subsection{Sampling and inference settings}

For each (model, dataset, variant) cell we draw $500$ examples from the
validation split using
\verb|random.Random(42).sample(range(len(split)), 500)|. The sampled
index list is built once per dataset and reused across models,
variants, and elicitation phrasings, so every cell is evaluated on the
same $500$ examples. All accuracy and ECE comparisons are paired at
the example level by construction.

All runs use a single NVIDIA L4 GPU, vLLM with
\verb|enforce_eager=True|, bfloat16 weights, fixed seed $42$, and
greedy decoding (\verb|temperature=0.0|, no nucleus or top-$k$
sampling). Generation budgets per call are: \verb|max_new_tokens=1|
with \verb|logprobs=200| for the dedicated label-extraction call
returning \verb|conf_token|; \verb|max_new_tokens=8| for the
verbal-confidence follow-up; \verb|max_new_tokens=32| for the
answer-scoring call, extended to $256$ on \texttt{format\_change} and
reduced to $4$ on the constrained-decoding repair path of
Appendix~\ref{app:finding1}. No stop sequences are used; generation
halts at EOS or the token budget. The software stack is Python 3.12.3,
vLLM 0.17.0, and PyTorch 2.10.0.

\subsection{Scoring, token probabilities, and verbal parsing}

\paragraph{Answer-letter scoring.} Accuracy is computed by
\verb|is_correct_mc|, which marks an example correct if the generation,
or its last non-empty line, begins with the gold choice letter. Token
confidence is computed separately via \verb|extract_pred_letter|, which
returns the first character of the first non-empty line if that
character is a valid choice letter (A--D for 4-choice, A--J for
10-choice MMLU-Pro); otherwise it reverse-scans lines for a standalone
choice letter; otherwise it falls through to the raw first character.
Whitespace, leading punctuation, and markdown bullet markers are
stripped before the first-character test. These are two different
definitions of the model's prediction; \verb|ECE_token| therefore pairs
confidence in the letter identified by \verb|extract_pred_letter| with
correctness as judged by \verb|is_correct_mc|. We report this because
the two rules diverge on chain-of-thought outputs. Every primary
tokenizer represents each single-letter option A--J as exactly one
token, so the top-$k$ logprob window returns letter probabilities
directly.

\paragraph{Token-confidence definitions.} Let $\ell^*$ denote the
predicted answer letter at the first generation step and $L$ the set
of valid choice-letter tokens observed in the returned top-$k$ logprob
window. We use two first-token confidence definitions:

\begin{align}
\hat{p}_{\mathrm{raw}}(\ell^*) &= P(\ell^*) \\
\hat{p}_{\mathrm{norm}}(\ell^*) &=
\frac{P(\ell^*)}{\sum_{\ell \in \mathcal{L}} P(\ell)}
\end{align}

The label-set-normalized form measures how probability is allocated
within the answer space and removes mass the model spends on
non-letter tokens (whitespace, punctuation, reasoning-word openings).
Both forms use first-token probabilities; neither measures the
probability of the final answer line a chain-of-thought model
eventually emits. ECE on either confidence uses equal-width binning
with $n_{\text{bins}}=10$. Main-text \verb|ECE_token| values use the
normalized form.

\paragraph{Verbal-confidence parser.} The parser accepts explicit
percentages (\texttt{75\%}, \texttt{75 percent}), decimal forms
(\texttt{0.75}, \texttt{.75}), and phrase forms
(\texttt{X out of 100}), in that order of precedence. Strict parses
only: the first well-formed value in $[0,1]$ for \texttt{decimal\_01},
or the first well-formed value in $[0\%,100\%]$ for
\texttt{percent\_0\_100}. Partial or multi-valued outputs do not parse.
Parsed values are clipped to $[0,1]$; parse failure returns
\texttt{None}, the example is excluded from \verb|ECE_verbal|, and the
example is counted as a non-parse for VPR.



\FloatBarrier
\section{Prompt templates and elicitation design}
\label{app:prompts}

\subsection{Prompt variants}

The five prompt variants are defined per task type
(\textit{classification}, \textit{reasoning}, \textit{QA}). For
classification and QA the \texttt{format\_change} variant modifies the
output-format instruction; for reasoning it appends a chain-of-thought
instruction. Full per-task YAML templates are in
\verb|src/prompts/{classification,reasoning,qa}_prompts.yaml|. We
reproduce the reasoning variants here because they drive Finding~1.

\paragraph{Reasoning prompt variants} (placeholders: \texttt{\{input\}}
$=$ question with lettered choice list; \texttt{\{label\_list\}} $=$
comma-separated labels; \texttt{\{fewshot\_examples\}} $=$ rendered
few-shot block):
\begin{itemize}
\item \texttt{surface\_paraphrase} (baseline):
``\textit{Answer the following multiple-choice question. \{input\}
Answer with only the letter of the correct option. Answer:}''
\item \texttt{instruction\_reorder} (instruction placed before
question): ``\textit{Choose the correct answer: \{label\_list\}.
Question: \{input\}. Correct answer letter:}''
\item \texttt{fewshot\_3} (three in-context examples injected):
``\textit{For each question, choose the correct answer from the
options provided. \{fewshot\_examples\} \{input\} Answer:}''
\item \texttt{format\_change} (chain-of-thought,
\verb|max_new_tokens=256|): ``\textit{Answer the following question.
Think step by step, then give the letter of the correct answer on the
last line. \{input\} Reasoning and answer:}''
\item \texttt{implicit\_framing} (no explicit task instruction):
``\textit{\{input\} The answer is:}''
\end{itemize}
Classification and QA variants follow the same five-failure-mode
structure with task-appropriate templates.

\subsection{Verbal-confidence elicitation prompts}

We use two structurally distinct elicitation suffixes appended to the
completed MCQ answer in a follow-up call. The \emph{primary} suffix
(\texttt{decimal\_01}) is:
``\textit{Now state your confidence in your answer above as a decimal
probability between 0.0 and 1.0. Reply with only the number, e.g.\
0.8.}''
The \emph{replication} suffix (\texttt{percent\_0\_100}) is:
``\textit{How confident are you in your answer? Respond with a single
percentage between 0\% and 100\% (e.g., 75\%).}''
The two phrasings differ on scale (decimal vs.\ percentage), mood
(declarative vs.\ interrogative), and expected output format (bare
number vs.\ number with \%). Each suffix is held fixed across all runs
for a given phrasing; variant perturbation applies to the MCQ prompt,
not to the elicitation. Primary numbers in the main text use
\texttt{decimal\_01}; \texttt{percent\_0\_100} is used for the
cross-phrasing robustness checks.

\FloatBarrier
\section{Additional evidence for Finding 1}
\label{app:finding1}

\subsection{ARC-Challenge evaluator repairs}

The ARC-Challenge collapse under \texttt{format\_change} reflects the
interaction of two pipeline choices: appending a chain-of-thought
instruction to the prompt, and the rule used to score accuracy.
Accuracy in our runs is scored by \verb|is_correct_mc|
(Appendix~\ref{app:repro}), which marks an example correct only if the
generation, or its last non-empty line, begins with the gold letter. On
\texttt{format\_change} cells the model typically opens with a reasoning
word (``The\dots'', ``Looking\dots'', ``Let\dots'', ``First\dots''), so
the first character is \texttt{T}, \texttt{L}, or \texttt{F}; the final
line is usually prose in which the answer letter is embedded rather
than a line beginning with that letter. The example is then scored
wrong regardless of what the model concludes.

Applying \verb|extract_pred_letter| to the same stored generations
instead, whose reverse scan searches lines for a standalone choice
letter, gives $0.472$--$0.712$ across the five primary models against a
non-chain-of-thought baseline of $0.734$--$0.890$ under that same rule,
a drop of $5.9$--$43.0\%$. No model falls below the $0.25$ random
baseline under this rule. The repairs described below are therefore
best read as recovering performance lost to the scoring rule used in
our runs, not as establishing a model-side ceiling.

\paragraph{Repair Path 1 -- regex re-parse.} An offline scorer applied
to the same per-example generation text. The regex extractor searches
for explicit answer-marker patterns in this order:
\verb|"Final answer: X"|, \verb|"the correct answer is X"|,
\verb|"Answer: X"|, and similar surface forms. If none matches, it
reverse-scans the final $150$ characters for a standalone choice
letter. If neither rule fires, it falls through to the first-character
rule. Path~1 needs only the stored CoT text and is therefore
applicable to any pipeline that retains generation outputs.

\paragraph{Repair Path 2 -- constrained-decoding pass.} An
inference-time repair: the same CoT generation is concatenated with
the literal string \verb|"\n\nFinal answer:"|, the resulting prompt is
fed back to vLLM under \verb|guided_choice = A/B/C/D|, and exactly one
letter is generated (\verb|max_new_tokens=4|, greedy). The constrained
vocabulary forces a well-formed label even on cells where the
underlying CoT body is malformed.

\paragraph{Stability of recoveries.} Across the five primary models,
fewer than $3\%$ of previously-correct first-character predictions
become wrong under either repair ($\leq 14$ of $500$ per model). The
two paths diverge substantively only on Qwen-2.5-7B, which emits
pathological repetition (\textit{``B. B. B. B \dots''}) and truncation
under the \verb|max_tokens=256| cap on a minority of cells: Path~1
then matches an early repeated letter rather than the model's
intended answer, while Path~2 ignores the malformed body and forces a
constrained letter from the appended re-prompt. Path~1's reduced
recovery on Qwen-2.5-7B is therefore a regex-fragility effect on a
specific output pattern, not a model-side ceiling. Per-model recovery
percentages are reported in Table~1 of the main paper.

\subsection{Calibration-normalization sensitivity}

Across the $125$ cells in the primary panel ($5$ instruct models
$\times$ $5$ datasets $\times$ $5$ variants), the per-cell mean
absolute difference between $\hat{p}_{\text{raw}}$- and
$\hat{p}_{\text{norm}}$-based ECE is $0.149\;[0.118, 0.184]$.
Table~\ref{tab:appC_normshift} disaggregates by prompt variant. The
shift is largest on \texttt{fewshot\_3} and \texttt{format\_change}
cells, where first-token mass is dispersed across in-context-example
letters or chain-of-thought reasoning words and the raw probability of
the answer letter is small even when the model concentrates
probability cleanly within the valid label set.

The values in this subsection pair confidence in the letter identified
by \verb|extract_pred_letter| with correctness as judged by
\verb|is_correct_mc| (Appendix~A.3). Recomputing with a single
prediction definition driving both quantities gives a per-cell mean
absolute shift of $0.114$ (median $0.017$), with $32.0\%$ of cells
shifting by more than $0.05$ and $18.4\%$ by more than $0.20$, and a
single-cell maximum of $0.894$ on Mistral-7B-Instruct's SST-2
\texttt{fewshot\_3} cell. The normalization effect is therefore
smaller than the figures reported here and concentrated in a heavy
tail rather than typical of most cells, but it does not vanish.

\begin{table}[ht!]
\centering
\caption{Per-variant ECE\_token shift under label-set normalization.
$n=25$ cells per variant; cell-level percentile-bootstrap CIs on the
mean $|\Delta|$. The all-cells row aggregates over the full
$125$-cell corpus. The single-cell maximum ($0.825$) occurs on
Mistral-7B-Instruct's SST-2 \texttt{format\_change} cell ($0.891$
unnormalized, $0.066$ normalized).}
\small
\begin{tabular}{lccccc}
\toprule
\textbf{Variant} & \textbf{Mean $|\Delta|$ [95\% CI]} & \textbf{Median} & \textbf{Max} & \textbf{$>0.05$} & \textbf{$>0.20$} \\
\midrule
\texttt{surface\_paraphrase}  & 0.135 [0.064, 0.217] & 0.028 & 0.596 & 40\% & 24\% \\
\texttt{instruction\_reorder} & 0.085 [0.050, 0.133] & 0.056 & 0.502 & 56\% & 4\%  \\
\texttt{fewshot\_3}           & \textbf{0.247 [0.169, 0.335]} & \textbf{0.204} & 0.655 & \textbf{72\%} & \textbf{52\%} \\
\texttt{format\_change}       & 0.173 [0.096, 0.270] & 0.100 & 0.825 & 56\% & 24\% \\
\texttt{implicit\_framing}    & 0.107 [0.056, 0.167] & 0.047 & 0.501 & 48\% & 16\% \\
\midrule
\textbf{All cells ($n=125$)}  & \textbf{0.149 [0.118, 0.184]} & \textbf{0.065} & \textbf{0.825} & \textbf{54.4\%} & \textbf{24.0\%} \\
\bottomrule
\end{tabular}
\label{tab:appC_normshift}
\end{table}

\subsection{Top-$k$ logprob coverage}

Under \verb|logprobs=200| the full label set is captured in $98.0\%$
of SST-2, $99.3\%$ of MNLI, $89.4\%$ of AG~News, $84.0\%$ of
ARC-Challenge, and $78.6\%$ of MMLU-Pro examples
(Table~\ref{tab:appC_coverage}). On cells where one or more letter
tokens fall outside the top-$200$ window, the uncovered letter carries
negligible mass: the per-cell mass-level coverage ratio exceeds $0.96$
on every dataset, so the normalized confidence and the label set used
for normalization are not materially distorted by truncation. Residual
truncation contribution to the reported per-variant $|\Delta|$ is
therefore second-order, and label-set normalization rather than top-$k$
truncation is the dominant effect.

\begin{table}[ht!]
\centering
\caption{Top-$200$ logprob window coverage for the valid-letter set.
``Cells with full coverage'' counts cells (out of $25$ per dataset)
for which $100\%$ of examples returned every valid letter.
``Example-level coverage'' is the per-dataset fraction of examples for which the full label set was captured.}
\small
\begin{tabular}{lcc}
\toprule
\textbf{Dataset} & \textbf{Cells with full coverage} & \textbf{Example-level coverage} \\
\midrule
SST-2          & 20 / 25 & 98.0\% \\
MNLI           & 22 / 25 & 99.3\% \\
AG~News        & 20 / 25 & 89.4\% \\
ARC-Challenge  & 16 / 25 & 84.0\% \\
MMLU-Pro       & 14 / 25 & 78.6\% \\
\bottomrule
\end{tabular}
\label{tab:appC_coverage}
\end{table}

\FloatBarrier
\section{Additional evidence for Finding 2}
\label{app:finding2}

\subsection{Verbal parse rate and threshold sensitivity}

VPR (Verbal Parse Rate) is the fraction of responses on a given
(model, dataset, variant) cell from which a numerical confidence can
be extracted under the parser of Appendix~\ref{app:prompts}. We use a
$\text{VPR} \geq 0.80$ inclusion threshold for \verb|ECE_verbal|
means; cells below threshold are excluded and footnoted, as including
them biases \verb|ECE_verbal| estimates because the parsed subset is
no longer representative of the cell's verbal-confidence distribution.

The $0.80$ threshold is tighter than dropping only zero-parse cells
and looser than requiring $\geq 0.95$. Findings~1--2 and the Mistral
cross-variant pattern do not depend on the specific value: at $0.70$,
$0.80$, $0.90$ the per-model \verb|overconf_vs_acc| means on MMLU-Pro
change by at most $0.020$, and the sign is strictly positive at every
threshold for every model (Table~\ref{tab:appD_threshold}).

\begin{table}[ht!]
\caption{Per-model \texttt{overconf\_vs\_acc} mean on MMLU-Pro under
three VPR thresholds (\texttt{decimal\_01}, valid variants:
\texttt{surface\_paraphrase}, \texttt{instruction\_reorder},
\texttt{implicit\_framing}). $n$ is the number of variants
contributing to each per-threshold mean.}
\centering
\small
\begin{tabular}{lccc}
\toprule
\textbf{Model} & \textbf{$\geq 0.70$ [$n$]} & \textbf{$\geq 0.80$ [$n$]} & \textbf{$\geq 0.90$ [$n$]} \\
\midrule
Llama-3.2-3B  & $+0.600$ [3] & $+0.600$ [3] & $+0.580$ [2] \\
Phi-4-mini    & $+0.614$ [3] & $+0.608$ [2] & $+0.608$ [2] \\
Gemma-3-4B    & $+0.784$ [3] & $+0.784$ [3] & $+0.784$ [3] \\
Mistral-7B    & $+0.707$ [2] & $+0.707$ [2] & $+0.707$ [2] \\
Qwen-2.5-7B   & $+0.673$ [3] & $+0.673$ [3] & $+0.673$ [3] \\
\bottomrule
\end{tabular}
\label{tab:appD_threshold}
\end{table}

\subsection{Mistral parse-rate matrix}

The main-text claim that Mistral's verbal-parse failure under
\texttt{surface\_paraphrase} is variant-specific---failing on every
dataset on that variant while clearing the $0.80$ threshold on the
majority of other-variant cells---is supported by the full
dataset~$\times$~variant matrix in Table~\ref{tab:appD_mistral_matrix}.

\begin{table}[ht!]
\centering
\caption{Mistral-7B-Instruct VPR by dataset $\times$ variant under
\texttt{decimal\_01}; four non-\texttt{format\_change} variants.
``$\times$'' marks cells below the $\text{VPR} \geq 0.80$ inclusion
threshold.}
\small
\begin{tabular}{lcccc}
\toprule
\textbf{Dataset} & \texttt{surface\_paraphrase} & \texttt{instruction\_reorder} & \texttt{fewshot\_3} & \texttt{implicit\_framing} \\
\midrule
SST-2          & 0.002\,$\times$ & 0.942 & 0.940 & 0.988 \\
MNLI           & 0.046\,$\times$ & 0.346\,$\times$ & 0.948 & 1.000 \\
AG~News        & 0.006\,$\times$ & 0.366\,$\times$ & 0.866 & 0.994 \\
ARC-Challenge  & 0.206\,$\times$ & 1.000 & 0.954 & 1.000 \\
MMLU-Pro       & 0.424\,$\times$ & 0.996 & 0.286\,$\times$ & 0.952 \\
\bottomrule
\end{tabular}
\label{tab:appD_mistral_matrix}
\end{table}

\subsection{Cross-phrasing robustness}

Under the alternative \texttt{percent\_0\_100} elicitation phrasing,
absolute VPR magnitudes on Mistral's \texttt{surface\_paraphrase}
cells rise substantially (Table~\ref{tab:appD_mistral_phrasing}). The
existence of below-threshold cells and their cross-dataset variation
pattern survive the phrasing change; the specific magnitudes do not.

\begin{table}[ht!]
\centering
\caption{Mistral-7B-Instruct VPR on \texttt{surface\_paraphrase} under
both elicitation phrasings; $n=500$ per cell; $95\%$
percentile-bootstrap CIs in brackets. Three of the five cells stay
below the $0.80$ threshold under \texttt{percent\_0\_100};
ARC-Challenge crosses threshold only under \texttt{percent\_0\_100}.}
\small
\begin{tabular}{lccc}
\toprule
\textbf{Dataset} & \textbf{VPR (\texttt{decimal\_01})} & \textbf{VPR (\texttt{percent\_0\_100})} & \textbf{$\Delta$} \\
\midrule
SST-2          & 0.002 [0.000, 0.006] & 0.324 [0.284, 0.364] & $+0.322$ \\
MNLI           & 0.046 [0.030, 0.064] & 0.134 [0.104, 0.164] & $+0.088$ \\
AG~News        & 0.006 [0.000, 0.014] & 0.100 [0.074, 0.126] & $+0.094$ \\
ARC-Challenge  & 0.206 [0.172, 0.242] & 0.776 [0.738, 0.814] & $+0.570$ \\
MMLU-Pro       & 0.424 [0.382, 0.470] & 0.642 [0.600, 0.684] & $+0.218$ \\
\bottomrule
\end{tabular}
\label{tab:appD_mistral_phrasing}
\end{table}

Applying the verbal-overconfidence analysis independently to each
phrasing run, the sign of \verb|overconf_vs_acc|,
\verb|overconf_vs_token|, and \verb|ECE_gap| is preserved on all five
primary models, and four of five shift by less than $0.10$ on
\verb|overconf_vs_acc| (Table~\ref{tab:appD_phrasing}). Llama-3.2-3B
is the exception: only one of three valid variants
(\texttt{instruction\_reorder}) passes $\text{VPR} \geq 0.80$ under
\texttt{percent\_0\_100}, leaving its per-model mean a single-cell
estimate. The full per-cell signed-overconfidence table is released
with the repository; every contributing cell on MMLU-Pro has a
strictly positive $95\%$ percentile-bootstrap interval on both
\verb|overconf_vs_acc| and \verb|overconf_vs_token|.

\begin{table}[ht!]
\centering
\caption{MMLU-Pro signed overconfidence under two elicitation
phrasings (valid variants $\cap$ $\text{VPR} \geq 0.80$, applied
independently to each run). $n$ values indicate the number of variants contributing to each per-model mean.}
\small
\begin{tabular}{lccccccc}
\toprule
\textbf{Model} & \textbf{verbal $-$ acc (dec)} & \textbf{(pct)} & \textbf{verbal $-$ tok (dec)} & \textbf{(pct)} & \textbf{ECE\_gap (dec)} & \textbf{(pct)} & \textbf{$n_{\text{dec}}/n_{\text{pct}}$} \\
\midrule
Llama-3.2-3B & $+0.600$ & $+0.316$ & $+0.604$ & $+0.371$ & $+0.576$ & $+0.437$ & 3 / 1 \\
Phi-4-mini   & $+0.608$ & $+0.700$ & $+0.466$ & $+0.553$ & $+0.470$ & $+0.552$ & 2 / 3 \\
Gemma-3-4B   & $+0.784$ & $+0.770$ & $+0.253$ & $+0.239$ & $+0.260$ & $+0.247$ & 3 / 3 \\
Mistral-7B   & $+0.707$ & $+0.702$ & $+0.490$ & $+0.486$ & $+0.490$ & $+0.487$ & 2 / 2 \\
Qwen-2.5-7B  & $+0.673$ & $+0.648$ & $+0.337$ & $+0.316$ & $+0.360$ & $+0.335$ & 3 / 3 \\
\bottomrule
\end{tabular}
\label{tab:appD_phrasing}
\end{table}

\subsection{Reliability diagrams}

\begin{figure}[ht!]
\centering
\includegraphics[width=\linewidth]{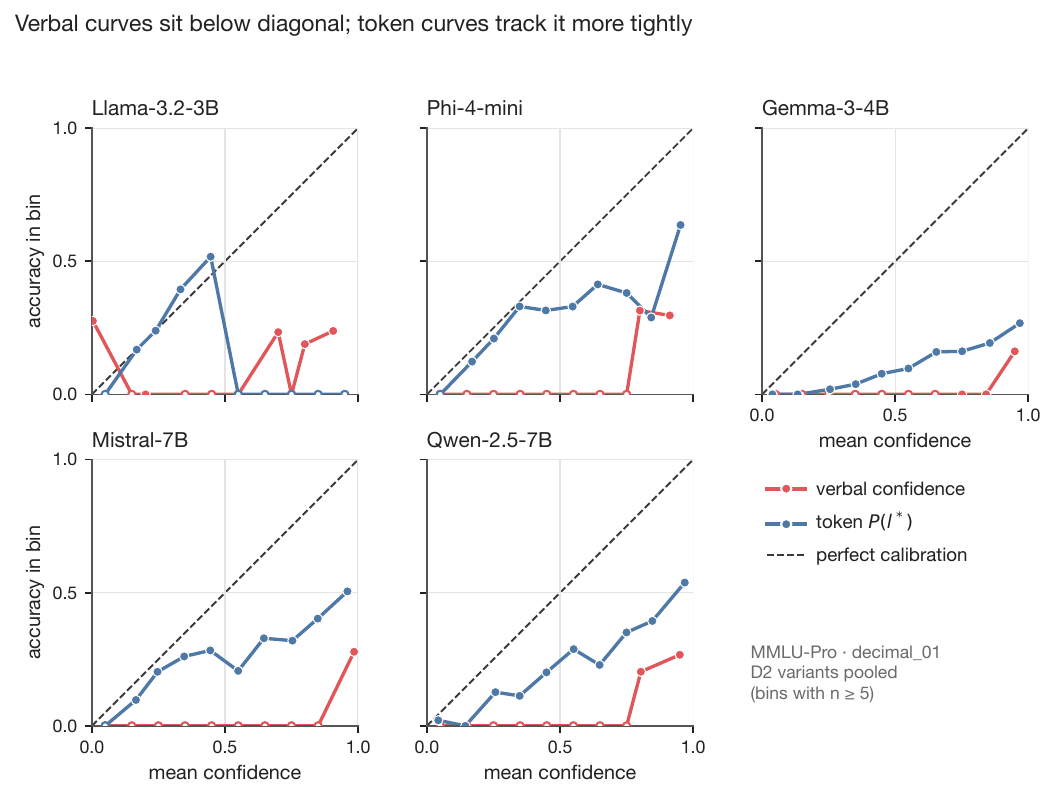}
\caption{MMLU-Pro reliability diagrams, one per primary model, under
\texttt{decimal\_01}. Verbal curve (red) and token curve (blue)
against the diagonal, pooled across variants with
$\text{VPR} \geq 0.80$.}
\label{fig:appD_reliability}
\end{figure}

Figure~\ref{fig:appD_reliability} gives per-primary-model reliability
diagrams on MMLU-Pro under \texttt{decimal\_01}, pooled across
variants with $\text{VPR} \geq 0.80$ (bin-weighted). Each panel
overlays the verbal curve and the token curve against the diagonal.
The verbal curve lies far below the diagonal in every panel.

\FloatBarrier
\section{Additional evidence for Finding 3}
\label{app:finding3}

\subsection{Size--spread correlations}

Finding~3's claim that parameter count does not order models by
prompt-perturbation spread rests on the five-benchmark Spearman
correlation panel below ($n=10$ instruct models per row).

\begin{table}[ht!]
\centering
\caption{Spearman $\rho$ between parameter count and
prompt-perturbation spread across the $10$-instruct-model sample.
None of the five correlations reaches $0.50$ in absolute value, and
all five $p$-values exceed $0.10$.}
\small
\begin{tabular}{lccc}
\toprule
\textbf{Dataset} & \textbf{Spearman $\rho$} & \textbf{$p$-value} & \textbf{$n$} \\
\midrule
SST-2          & $+0.261$ & 0.466 & 10 \\
MNLI           & $+0.474$ & 0.166 & 10 \\
AG~News        & $-0.244$ & 0.497 & 10 \\
ARC-Challenge  & $+0.015$ & 0.967 & 10 \\
MMLU-Pro       & $+0.304$ & 0.393 & 10 \\
\bottomrule
\end{tabular}
\label{tab:appE_spearman}
\end{table}

The within-family clustering noted in Finding~3 (both Qwen models
among the most robust on SST-2; both Gemma models among the most
brittle on ARC-Challenge) is a descriptive observation on a
$10$-model convenience sample. We do not fit a family-level model and
do not separate family effects from training-recipe effects.

\subsection{Bootstrap procedure}

All confidence intervals reported in the paper are $95\%$ percentile
bootstrap intervals with $n=1000$ resamples and fixed seed $42$.
Resampling units depend on the metric:
\begin{itemize}
\item \textbf{Calibration normalization shift:} cell-level resampling
of the $25$ per-variant $|\Delta|$ values (or the $125$ values for the
all-cells row); statistic $=$ \verb|numpy.mean|.
\item \textbf{ARC-Challenge paired drop:} paired example-level
resampling of (correct under \texttt{surface\_paraphrase}, correct
under \texttt{format\_change}) tuples within a (model, dataset) cell;
statistic $=$
$\text{acc}_{\text{surface}} - \text{acc}_{\text{format\_change}}$.
\item \textbf{Per-cell ECE\_gap:} example-level resampling within each
cell that passes $\text{VPR} \geq 0.80$, pooled across variants for
the same model on MMLU-Pro; statistic $=$
$\text{ECE}(\text{conf}_{\text{verbal}}) - \text{ECE}(\text{conf}_{\text{token}})$
evaluated with $n_{\text{bins}}=10$. Unparseable rows are dropped
within-bin during ECE computation.
\item \textbf{Per-cell signed overconfidence:} paired example-level
resampling of (verbal-confidence, token-confidence, accuracy) triples
on the rows where verbal confidence parses; statistic $=$
$\overline{\text{conf}_{\text{verbal}}} - \text{accuracy}$ and
$\overline{\text{conf}_{\text{verbal}}} - \overline{\text{conf}_{\text{token}}}$
on the resample.
\item \textbf{Per-cell VPR:} example-level resampling of binary
parseable indicators within a single (model, dataset, variant) cell;
statistic $=$ \verb|mean(parseable)|.
\item \textbf{Prompt-perturbation spread:} example-level resampling
per (model, dataset); statistic $=$ \verb|max - min| over the four
non-\texttt{format\_change} per-variant accuracies computed on the
resample.
\end{itemize}
Cross-phrasing comparisons are bootstrapped on the two phrasing runs
\emph{independently}, not paired across phrasings, because the
example-level verbal responses differ by construction.

The bootstrap quantifies sampling variability in the per-cell or
per-aggregate estimate given the corpus we use. It does not bound
across-corpus generalization: it does not cover how the same
statistic would behave on a different sample of models, datasets, or
elicitation-prompt families.

\end{document}